\documentclass{llncs}
  
\usepackage[utf8]{inputenc} 
\usepackage{graphicx}
\graphicspath{ {images/} }
\usepackage{amssymb}
\usepackage{amsmath}
\usepackage{url}

\begin{document}
%\title{Exploit the Potential of the Group: \\ Putting Humans in the Dedicated Collaborative Interactive Learning Loop}
\title{Smart Device based Initial Movement Detection of Cyclists using Convolutional Neuronal Networks}
% \subtitle{}
\author{Jan Schneegans and Maarten Bieshaar}
\institute{Intelligent Embedded Systems Lab, University of Kassel, Germany\\
\email{\{j.schneegans, mbieshaar\}@uni-kassel.de}}

\maketitle

\begin{abstract}
For future traffic scenarios, we envision interconnected traffic participants, 
who exchange information about their current state, e.g., position, 
their predicted intentions, allowing to act in a cooperative 
manner. Vulnerable road users (VRUs), e.g., pedestrians and cyclists, will be equipped with smart device that can be used to detect their intentions and transmit these detected 
intention to approaching cars such that their drivers can be warned.
In this article, we focus on detecting the initial movement of cyclist using 
smart devices. Smart devices provide the necessary sensors, namely accelerometer and gyroscope, and therefore pose an excellent instrument to detect movement transitions (e.g., waiting to moving) fast. Convolutional Neural Networks prove to be the state-of-the-art solution for many problems with an ever increasing range of applications. 
Therefore, we model the initial movement detection as a classification problem.
In terms of \textit{Organic Computing (OC)} it be seen as a step towards self-awareness and self-adaptation.
We apply residual network architectures to the task of detecting 
the initial starting movement of cyclists.
\end{abstract}


\section{Introduction}
\label{js:sec:introduction}
We envision future mixed traffic scenarios~\cite{mbBRZ+17}, involving automated cars, trucks, sensor-equipped infrastructure, and other road users equipped with smart devices and other wearables which are interconnected by means of an ad hoc network and cooperate on different levels, e.g., on the level detected intentions.  Vulnerable road users (VRUs) will continue to take part in future urban traffic, recognizing them and detecting their intentions poses an important goal towards a safer traffic environment. Many vehicles already incorporate a range of sensors, but they still rely on an unhindered line of sight. The adoption of smart devices implements an alternative considering almost everyone has one with them while taking part/participating in traffic. It comes pre-equipped with the necessary hardware like accelerometer, gyroscope and magnetic field sensors and in near future will be capable to connect with its surrounding via ad hoc networks. GPS sensors enable only a rough localization and do not offer the sampling rates 
to detect fast movement changes. System based on inertial data allow for a fast detection of changes in motion. In this article we focus on a reliable detection of starting movement transitions, i.e., initial movement detection.
On the one hand side, the detector has to be robust, i.e., avoid false positive detections, and on the other hand side it has to be fast.
As we have shown in~\cite{mbBZD+17}, the initial movement detection can greatly support the VRU's trajectory forecast.

In terms of organic computing, the system requires self-reflection capabilities in order to adapt subsequent algorithms, e.g., 
for trajectory forecasting. 
In this article an approach to smart devices initial movement 
detection is presented, aiming to bring self-reflection capabilities and context awareness to smart devices in domain of VRU intention detection.

\subsection{Main Contributions and Outline}
\label{js:sec:outline}
The approach presented in this article aims to detect starting movements of cyclists fast and yet robust, i.e., with few false positive detections. The approach is based on an convolutional neural networks trained on the inertial data (accelerometer and gyroscope) obtained from the smart device worn by the cyclist.
\\

In Section~\ref{js:sec:related_work} the related work is reviewed. Section~\ref{js:sec:methodology} presents our approach including the preprocessing pipeline and the applied deep learning methods. Following a description of the data acquisition and evaluation methodology in Section~\ref{js:sec:data}, Section~\ref{js:sec:experimental_results} demonstrates the experimental results. Section \ref{js:sec:Conclusion} concludes the results and provides further insight into future work.

\section{Related Work}
\label{js:sec:related_work}
Previous work focused on the detection of human activities for various applications ranging from elder care, motion types like standing, walking up or down stairs etc., or even footstep detection \cite{singh2017}. Employing Deep Convolutional Neural Networks (CNNs), rather than handcraft features, appears to pose a viable option based on the success of this emerging technology. Applying machine learning to smart phone data was used for example by Jiang and Yin \cite{jiang2015} in order to perform human activity recognition. Their data is given by accelerometer and gyroscope of wearable devices. Deep Convolutional Neural Networks enable them to automatically learn the optimal features from activity images for the activity recognition task, providing state of the art results on data from wearable sensors. The key feature of their success relies on activity images created out of the Fourier Transform and permutation to enable the model to learn features connecting information contained in multiple tracks of data. The learned filters in a convolutional network thereby substitute handcrafted features which were in the past usually extracted independently from multiple time series sensor signals. This indicates the possibilities to detect further motion primitives like starting and stopping motions in a traffic environment.

Bieshaar et al.~\cite{mbBZH+18} envision a cooperative traffic environment where each individual is connected to the surrounding traffic participants and infrastructure, while maintaining a model of the present situation. They propose a system involving video data from cameras at intersections and smartphone data via a boosted stacking ensemble. This combination leads to a robust classification, although being a complex procedure. Their approach was able to detect 99\% of the starting movements within 0.24 ms after the first movement of the bicycle's back wheel. The system is comprised of a 3D-CNN for the image processing and a smart device-based starting movement detector consisting of an extreme gradient boosting classifier (XGBoost).

Working on a similar topic, Zernetsch et al. \cite{zernetsch2018} also operate on camera data but this time evaluate based on Motion History Images input into a Residual Network (ResNet) and a Support Vector Machine (SVM). Both are capable of a correct classification, whereby the ResNet approach is more robust and accurate than the SVM. In another article \cite{zernetsch2018_2}, they built a forecasting model predicting the location of vulnerable road users for the next 2.5\,s. The experiments include two methods: a physical model of a cyclist and a polynomial leas-squares approximation in combination with a multilayer perceptron. Both were compared to a Kahlman Filter approach and improved on its results. The physical model achieved a 27\% more accurate positional prediction, while the least-squares approximation attained a 34\% increase in accuracy for the starting and stopping motion of a cyclist with otherwise similar outcome as the first approach.

\section{Methodology}
\label{js:sec:methodology}
\begin{figure}
	\centering
	\includegraphics[width=\textwidth]{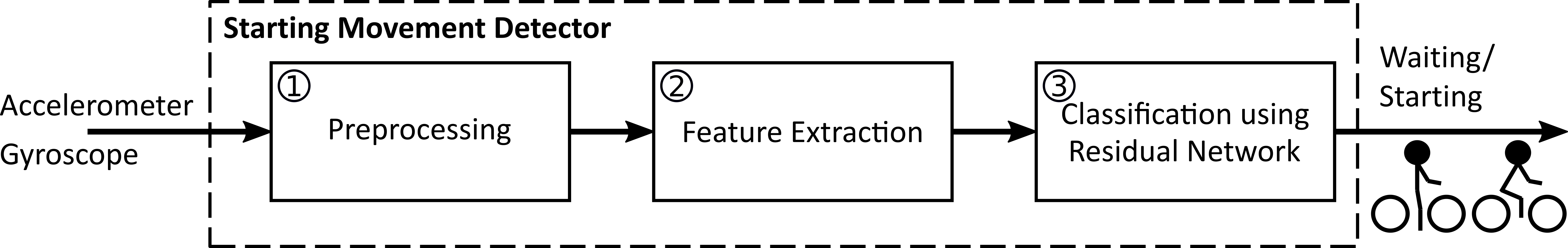}
	\caption[Processing pipeline consisting of preprocessing and feature extraction, feature selection, and classification.]
	{Processing pipeline consisting of three stages: preprocessing and feature extraction, feature selection, and classification using convolutional neural network taking the form of a residual neural network.}
	\label{fig:pipeline}
\end{figure}

Our approach aims to detect the movement transition between waiting and moving (i.e., starting) of cyclists as early as possible using a smart device. The starting movement detector is realized by means of a HAR pipeline~\cite{Bulling2014THA}. A schematic of the starting movement detector consisting of three stages is depicted in Fig. \ref{fig:pipeline}. The cyclists starting movement is modeled as a classification problem of the classes \textit{waiting}, \textit{starting} and \textit{moving}. Inertial data, i.e., accelerometer and gyroscope, is used as input for the starting movement detector. The data is preprocessed (i.e., transformed into a device attitude invariant representation) followed by a features extraction. Finally, the detection is realized by means of a deep learning based classifier taking the form of a residual neural network.

\subsection{Preprocessing and Feature Extraction}
\label{sec:preprocessing}
The first modules of the starting movement detector consist of the data preprocessing and feature extraction. We compare four different sets of features.

\begin{figure}
	\centering
	\includegraphics[width=\textwidth]{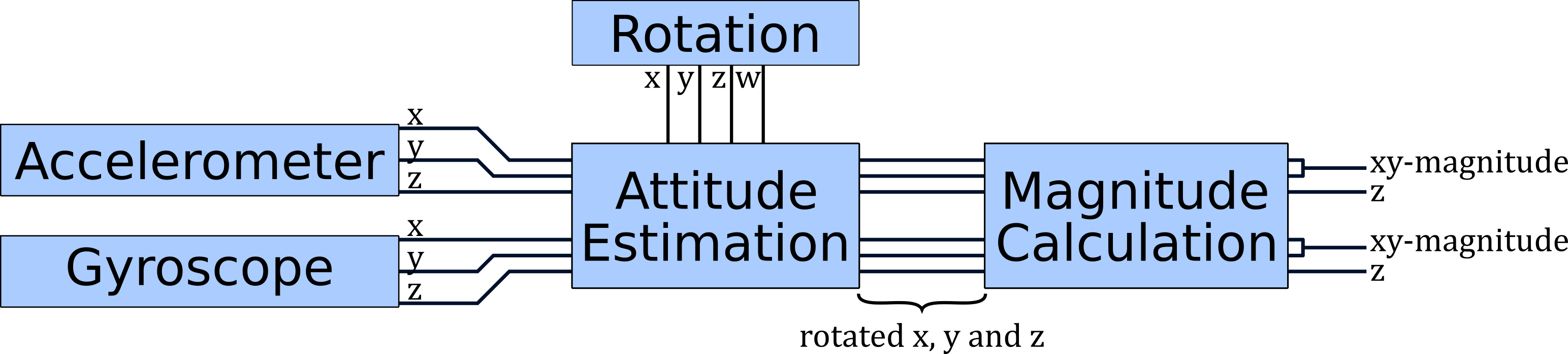}
	\caption[Two-step pre-processing of the raw sensor signal; rotation and magnitude calculation.]
	{Two-step pre-processing of the raw sensor signal. First attitude information is used to rotate x, y and z value of the accelerometer and gyroscope, then the magnitude of x and y is calculated.}
	\label{fig:preprocess_pipeline}
\end{figure}

The raw data is provided by the accelerometer and gyroscope. Before feeding it into the model, the x, y and z axis are rotated to lie flat on the earth, i.e., the z-axis pointing towards the sky. This is done with the help of rotational information provided by a magnetometer. To avoid the tedious estimation of device to VRU transformation, we consider the magnitude of the x and y axis. This leaves four features to be used: the gravity adjusted magnitude and z values of the accelerometer and gyroscope. This comprises the first set of features. A schematic of this procedure is depicted in Fig. \ref{fig:preprocess_pipeline}.

Additionally, the Fourier Transform up to the fifth degree was constructed based on a sliding window segmentation of the preprocessed data. It provides coefficients and phase over windows of size 640\,ms and 5120\,ms. This comprises the second set of features.

\begin{figure}
	\centering
	\includegraphics{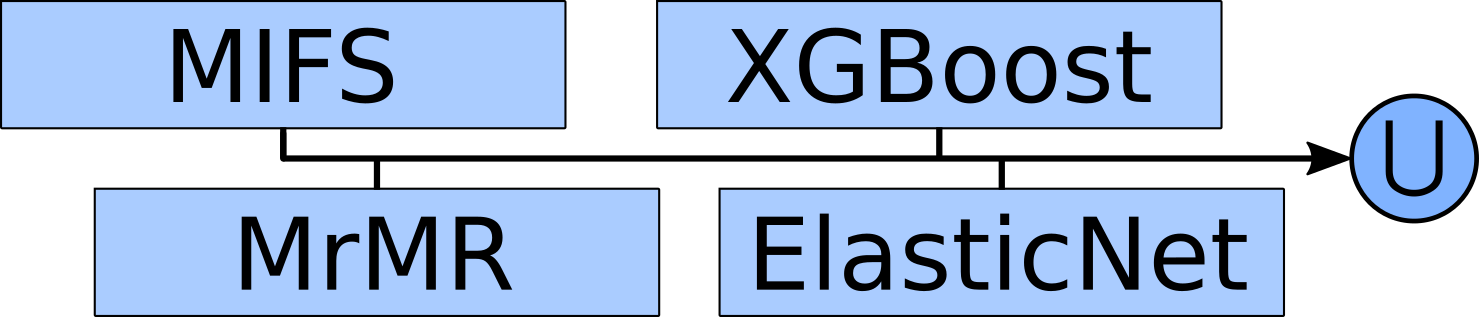}
	\caption[The filter pipeline reducing the number of selected features.]{The filter approach pipeline reducing the selected features. Four classifiers select the 10 best features and thereafter a union produces the final features set.}
	\label{fig:feature_selection}
\end{figure}

The preprocessed data was further handled to generate over 150 features, including the mean, variance, energy, min, max and polynomial characteristics over window sizes of 100\,ms, 500\,ms, 1000\,ms and 2000\,ms. Those features (including the Fourier coefficients) were sieved through a filter approach consisting of four procedures depicted in Fig. \ref{fig:feature_selection} that selects the ten most indicative features of each cross-validation fold separately. To obtain a large diversity concerning the selected features four complementary types of filters were picked. Two feature selection techniques: minimum redundancy maximum relevance (MrMR), mutual information (MIFS); and two model based methods applying an elastic net (ElasticNet) and extreme gradient boosted trees (XGBoost). An union set of those selected features was then utilized, amounting to roughly 19 features per fold. Throughout this text, this third set of features is referred to as filter features.

Moreover, the Fourier coefficients and phases were evaluated for each sensor (accelerometer, gyroscope) and window (5120\,ms, 640\,ms) separately. Also a combination of the filter features and the Fourier phase was tested to see if further improvements could be achieved. They were therefore simply concatenated creating the largest of input spaces containing 31 features and comprising the fourth set of features.

\subsection{Classification using Residual Networks}
\label{sec:classification}
The last stage of the starting movement detector realized by mean of deep learning, i.e., a residual neural network, which is a special kind of convolutional neural network. Deep learning has been recently proven to be extremely successful in various domains and has lead to an increased usage of machine learning techniques in a multitude of fields.

Convolutional neural networks (CNN) have already been applied to many practical tasks and have risen in popularity after achieving superhuman accuracy on image classification tasks \cite{ILSVRC15}. They work by sequentially applying convolutions over an input space and are trained via back-propagation of error gradients. Their input space typically takes the form of a 2D or 3D image with depth greater or equal to one. Stacking multiple layers of convolution helps to identify broader features and covers more of the input space the deeper the network grows, enabling the model to learn abstract characteristics regarding the data.

\subsubsection{Residual Networks}
\label{js:sec:Residual_Networks}
In standard convolutional networks, a layer is only connected to the preceding and proceeding layers limiting the information visible to deeper layers to activations of the preceding layer, i.e. only features capture by previous layer are attainable. An architecture getting around this constraint is the Residual Network (ResNet) proposed by He et al. \cite{he2015}. A network comprised of this ResNet architecture won the 1st place of the ImageNet Large Scale Visual Recognition Competition \cite{ILSVRC15} in 2015. It was originally designed to conquer the problem of degradation in very deep networks. When deeper networks start converging accuracy gets saturated and then degrades rapidly with increasing network depth. To resolve this issue, the authors introduced shortcut connections skipping convolutional layers and therefore providing deeper layers with the identity of the original input.
Instead of learning a direct mapping of $H(x): x \rightarrow y$, a residual function $F \left(x \right) = H\left(x\right) - x$ is defined. 
It can be reformed into $H(x) = F(x) + x$, where $F(x)$ and $x$ represent the non-linear layers, e.g., convolution, activation or regularization layers, and the identity function respectively.
According to the findings of He et al. the residual function $F(x)$ is easier to optimize than $H(x)$. For example when trying to learn the identity function, learning to push $F(x)$ to zero can be easily achieved using a stack of non-linear convolutional layers.

\begin{figure}[h]
	\centering
	\includegraphics[scale=0.55]{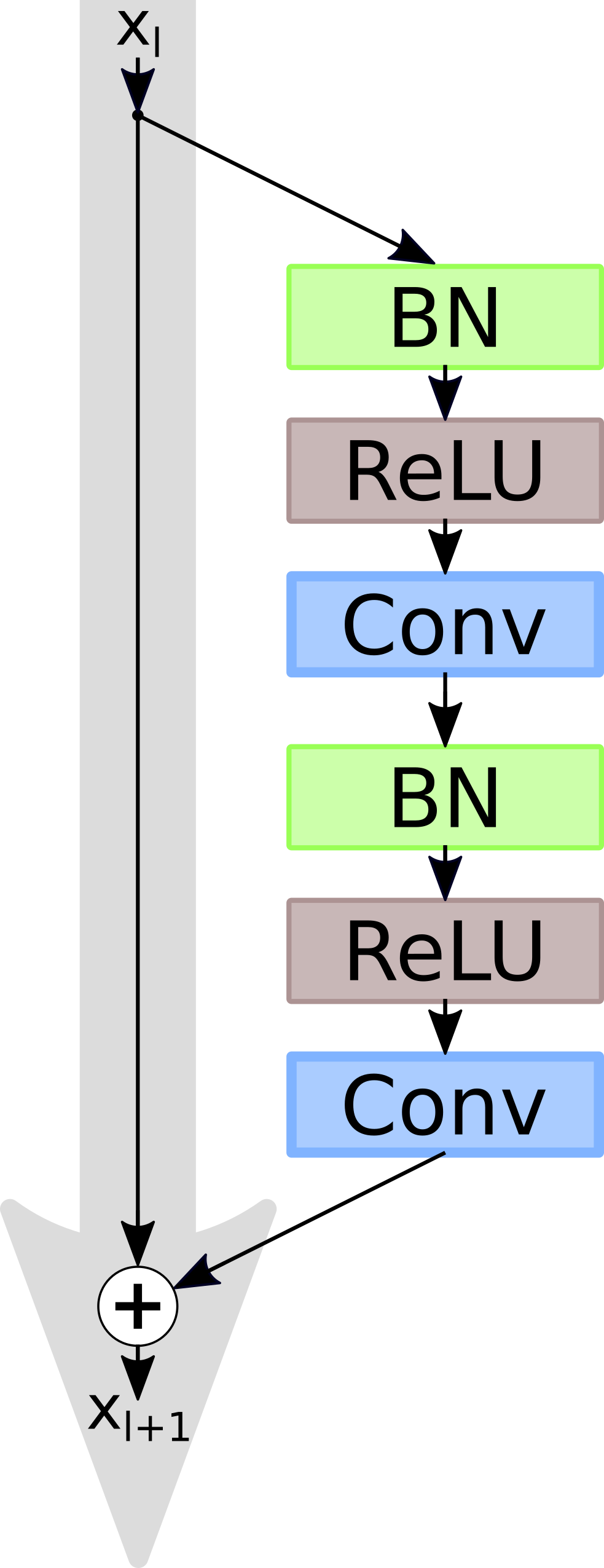}
	\caption[Schematic of a residual unit composed of the residual function $F(x)$ and the identity shortcut $x$.]
	{Schematic of a residual unit composed of the identity shortcut $x$ (left) and the residual function $F(x)$ (right) transforming the input from the previous layer $x_l$ to the input of the next layer $x_{l+1}$.}
	\label{fig:resunit}
\end{figure}

\paragraph{Residual Unit}
\label{js:sec:Residual_Unit}
The residual unit is a computational unit composing the identity shortcut $x$ and the residual function $F(x)$, which itself is composed of convolutional, activation and batch normalization layers \cite{ioffe2015}. He et al. strongly suggest to apply them in the order of batch normalization, activation function followed by the convolution. The best position of the shortcut is around two sets of layers ordered as just mentioned \cite{he2016}. A schematic of a residual unit is depicted in Fig \ref{fig:resunit}.

\subsubsection{Regularization}
\label{js:sec:Regularization}
Regularization methods aim at decreasing the validation and test error, thus increasing the generalization abilities of machine learning models. In the following the regularization schemes adopted in this work will be described.

\paragraph{Spatial Dropout}
\label{js:sec:Spatial_Dropout}
Spatial dropout is a special form of dropout being used in convolutional networks, motivating the learning of more independent features. In contrast to normal dropout, where single neurons are excluded, whole filters are set to zero. This mimics the original intent of omitting complete and randomly selected features during training more closely than just discarding parts of it. Although spatial dropout increases the temporal requirements while training, it encourages the learning of more robust features, consequently leading to a smaller validation and test error, i.e., better generalization ability.

\paragraph{Batch Normalization}
\label{js:sec:Batch_Normalization}
To speed up the learning process and increase the generalization ability, batch normalization can be applied. It is implemented by adding special layers before the convolutions, which normalizes the input signal of those layers according to the batch statistics, namely the mean and variance of the current mini-batch \cite{ioffe2015}.
This works because it minimizes the amount of covariance shift of the hidden units. As an example, consider a network trained on black cat images: When applying this network on colored cats, it will perform worse, because while it still depicts a cat, the distribution of color values is different. In other words, if an algorithm learned some X to Y mapping, and if the distribution of X changes, then we might need to retrain the learning algorithm by trying to align the distribution of X with the distribution of Y. Batch normalization is favoring a more stable learning of features by normalizing the inputs to deep hidden layers reducing the effect of covariance shift, because the mean and variance remain similar during training \cite{ng2017}.

Additional benefits of batch normalization include being able to use higher learning rates, since the input values to the convolutional layers are restricted from being very high or very low. It also adds a regularization effect by adding noise to the activations (due to the statistics being calculated only on the current mini-batch).

\subsubsection{Architecture}
\label{js:sec:Architecture}
\begin{figure}[h]
	\centering
	\includegraphics[scale=0.5]{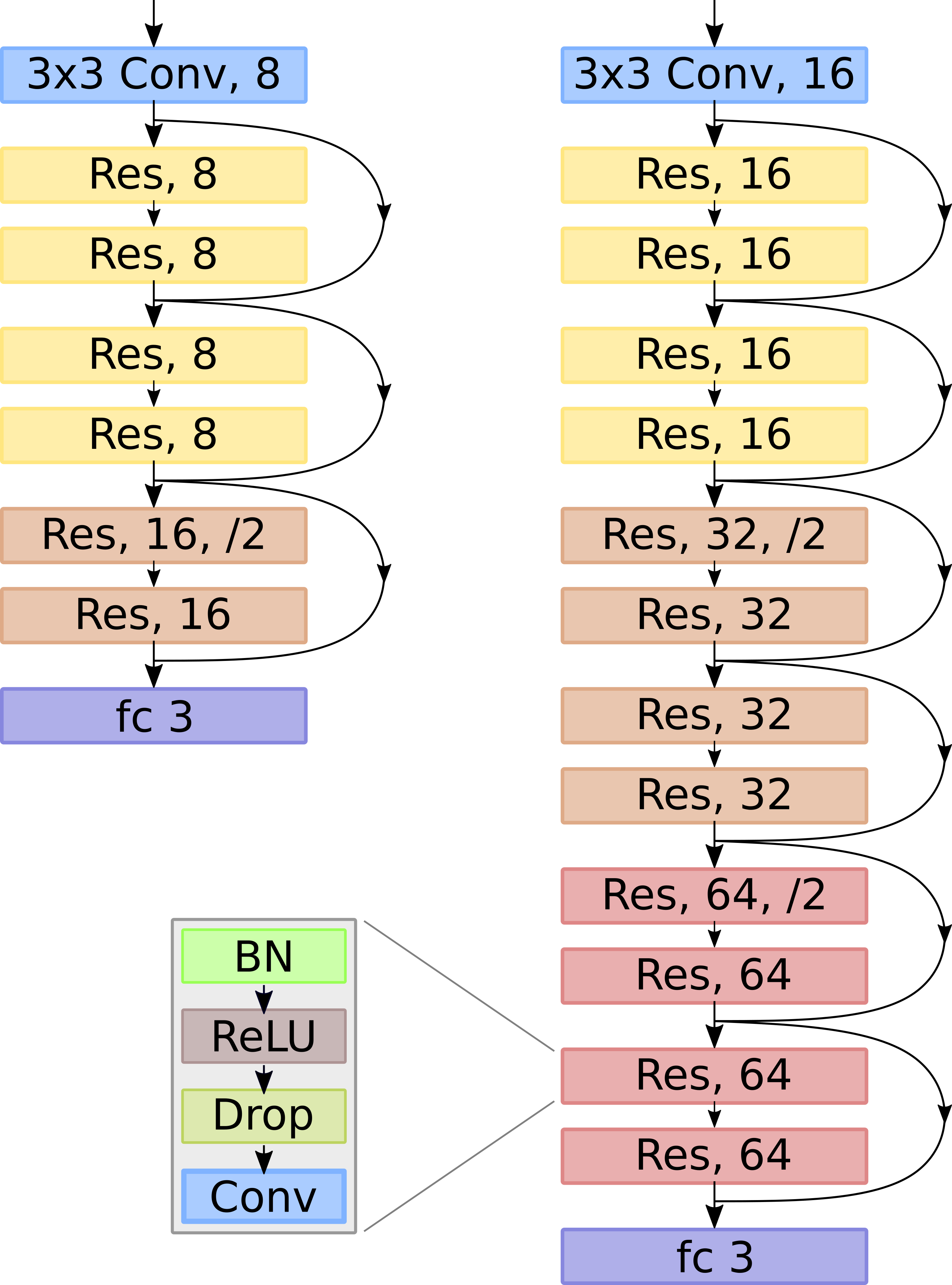}
	\caption[Architecture of the small eight layer and large 14 layer networks, with close up of a residual unit.] {Architecture of the networks used: (top-left) the small network consisting of eight layers and starting with eight filters. (right) the larger network consisting of 14 layers starting with 16 filters. The shortcut connection each encompasses two blocks of batch normalization, activation function (Rectified Linear Unit), spatial dropout and convolution as depicted in the bottom-left.}
	\label{fig:architecture}
\end{figure}

Two basic architectures of the ResNet were examined during the experiments. They are depicted in Fig \ref{fig:architecture}. The smaller one (Fig. \ref{fig:architecture}, top left) consists of two arrays of residual units, preceded by a first simple convolutional layer and followed by a fully connected layer to reduce the spatial dimensions to the desired output size of three classes. The first array holds two residual units and the seconds holds one. Softmax is applied to obtain values in the interval of $[0, 1]$. The network starts off with eight filters and doubles the amount for the second array of residual units, in this instance also the spatial dimension are halved. In this case, an additional convolution takes place in the shortcut to fit the passing identity to the new dimensions.

The larger network follows (Fig. \ref{fig:architecture}, right) the same architectural approach but with three arrays holding two residual units each and resulting in 14 layers also counting the first convolutional and last fully connected layers. The first layer and array begin with 16 filters, doubling the number of filters and reducing the spatial dimensions by one half every array, just as in the smaller model.

Both networks were trained to optimize the weighted cross-entropy applying the Adam-Algorithm. As indicated earlier, spatial dropout and batch normalization were applied. In this project the residual unit was structured as explained in Section \ref{sec:Residual_Unit} with the addition of a dropout layer right before each convolution (excluding the first convolutional layer), as can be seen in the lower left of Fig. \ref{fig:architecture}. The convolutions were carried out with 3x3 filters, stride set to one and zero padding.

\section{Data Acquisition and Evaluation}
\label{js:sec:data}

\subsection{Data Acquisition}
\label{js:sec:data_acqu}
The data of 49 test persons was gathered during a week long experiment at a public intersection incorporating bicycle lanes as well as walkways. It consists of 84 start scenes and the ground truth was generated through a wide angle stereo camera system and visual inspection~\cite{mbBZH+18}. The cyclist were instructed to cross the road or drive alongside it, adhering to the traffic rules. Each carried a smart phone (Samsung Galaxy S6) in the right trouser pocked.

\subsection{Evaluation}
\begin{figure}
	\centering
	\includegraphics[scale=0.2]{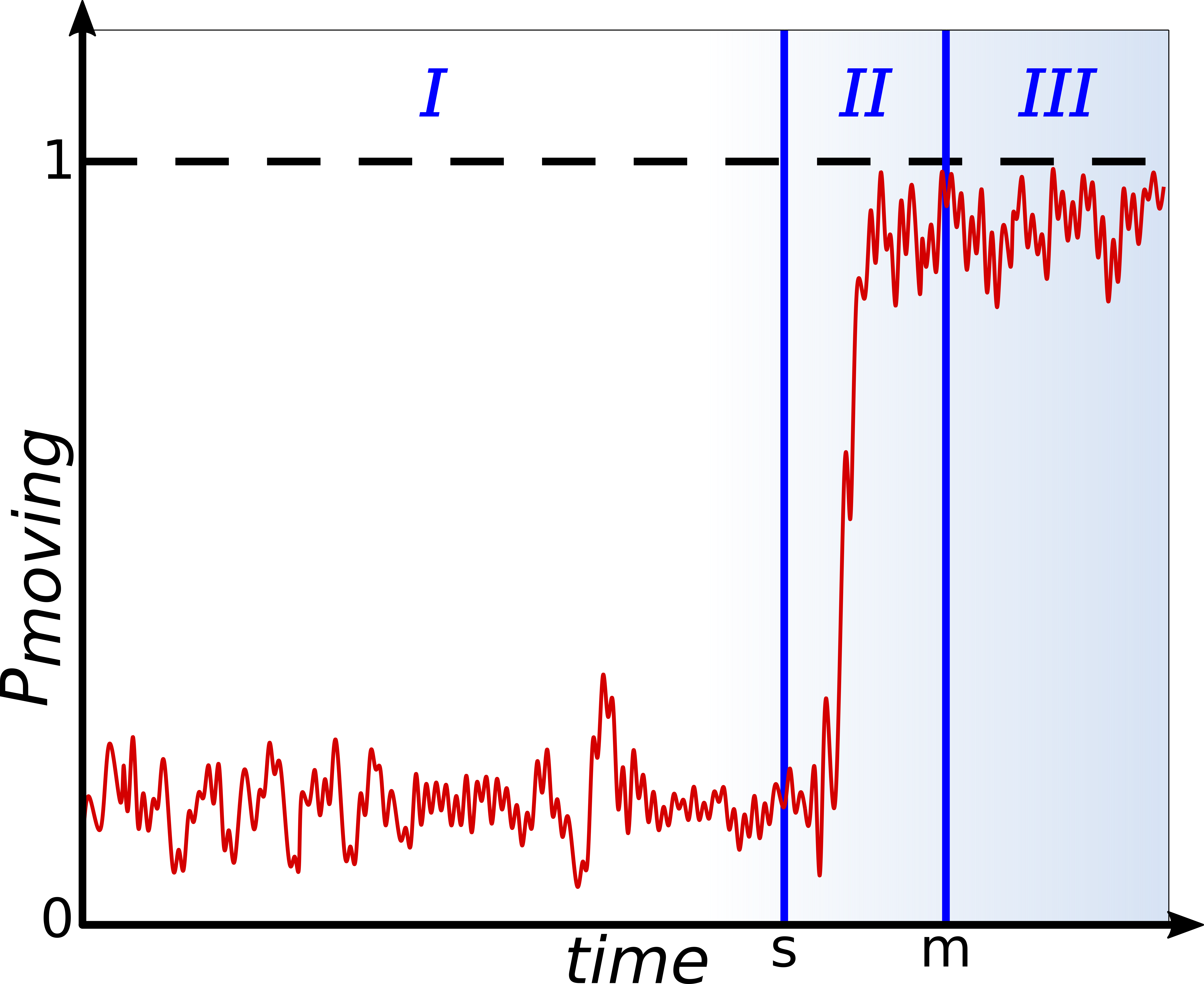}
	\caption[The three classes making up the labeling and separation of a time series with example probabilties for the third class.]
	{In red: example probabilities for the \textit{moving} class of a starting prediction. Area $I$ is the \textit{waiting} class of low activity; $II$ denotes the \textit{starting} class meaning the initial motion leading to the start, beginning at time $s$; $III$ is the \textit{moving} class labeled after the first movement of the bicycles back wheel, marked as $m$.}
	\label{fig:classes}
\end{figure}

We model the starting movement detection problem as a classification problem. It consists of three classes indicating \textit{waiting}, \textit{starting} and \textit{moving}, which split a time series in three parts. This separation is depicted in Fig. \ref{fig:classes}. The most common label is \textit{waiting} ($I$) and describes an interval of low movement and acceleration, while the cyclist is waiting, e.g., at a traffic light. An initial movement that leads to a start, including the starting motion up to the \textit{moving} phase is labeled as \textit{starting} ($II$). Thereby, \textit{moving} ($III$) is defined as the first movement of the back wheel followed by an acceleration and driving.

In order to calculate scores on the validation and test data the $F_1$ score was used. It is calculated from the precision and recall, which in turn are calculated from the true-positive, true-negative, false-positive and false-negative values defined as follows: if the prediction of \textit{moving} is after the true \textit{starting} label, it is classified as a true-positive. This means that a classification of a start is still classified as a true positive even if it is detected before the \textit{moving} label but during the initial movement of the test person intending to start. Else if it is too early, it is classified as a false-positive. True-negatives normally do not appear since every time series contains \textit{starting} label. When the predictor would not output the \textit{starting} class at any time step, it is classified as a false-negative.

The delay of the starting prediction is determined only for true positive classifications, as otherwise false positives would greatly reduce the average measured delay. It is calculated relative to the first true \textit{moving} label. Negative values are possible, because a classification during the \textit{starting} phase is still considered a true-positive.

%To assess the different models in addition to the test scores, plots were generated depicting an overview of the $F_1$-score, precision and delay for each model. Those statistics were calculated over the concatenation of each test folds results. Furthermore, histograms over the distribution of the delays were created and processed using a kernel density estimation. Having plots and statistics of every model separately is good but tedious to examine, therefore a graph displaying the Pareto optimality regarding the $F_1$-score and the delay was created containing all examined models.

\section{Experimental Results}
\label{js:sec:experimental_results}
The creation and evaluation of models for the starting intention detection is separated into two parts: selecting the input data and corresponding features before optimizing the model architecture.

\subsection{Feature Selection}
\label{js:sec:Feature_Selection}
First, the feature selection was conducted. The results turned out as expected: a smaller input window leads to smaller delays, a larger input window leads to larger delays. The convolutions themselves add to the delay as they cover multiple time steps on higher level learned features. The filter features lead to higher delays compared the raw features, due to the mean, variance etc. being calculated over windows as well. Nonetheless, they are more robust and provide a more stable score over different thresholds applied on the final softmax output.

The overall best scores were achieved by the filter features. On an input window of 100 time steps it reaches an $F_1$-score of 95\% and has a mean delay of 861\,ms. With an input window of 50 time steps it has a slightly faster mean delay of 783\,ms.
The raw features end up in the middle field, yet, not as good as the filter features with an input window of 10 time steps. With an input window of 100 time steps their highest $F_1$-score is 76\% with a very low mean delay of 257\,ms.

Combining the filter features with the phase of a discrete Fourier Transformation slightly reduces the scores compared to only filter features. Supposedly because the network is too small to deal with so many features. This assumption is strengthened considering that the same features are contained with addition of the phase.

Using the discrete Fourier Transform coefficients and phases thereof do not provide sufficient information for a starting intention detection. The highest $F_1$-score of 63\% was reached by the Fourier coefficients of the Accelerometer over a window of 5120\,ms with an input window of 100 time steps. All other Fourier features are worse, especially the phase reaching only a maximum $F_1$-score of 32\%.

\subsection{Architecture Optimization}
\label{js:sec:Architecture_Optimization}

\begin{figure}[h]
	\centering
	\includegraphics[width=\textwidth]{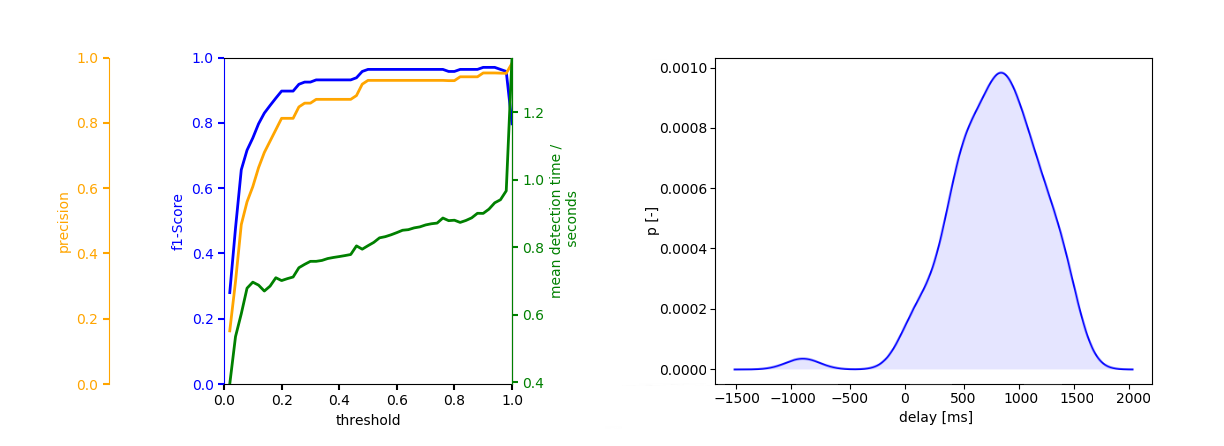}
	\caption[Classification results and Kernel Density Estimation of the delay of the best performing model]
	{Classification results (left) and Kernel Density Estimation of the delay (right) of the best performing model, which processes the filter features, has an input window of 50 time steps and keep probability of 0.6. Most classification occur between 0.5\,s and 1\,s after the \textit{moving} label. Noticeably, some starting detection happen almost 1\,s before the bicycle moves.}
	\label{fig:filter_06_results}
\end{figure}

For the architecture optimization, the filter features with an input window of 50 time steps was picked as a compromise between the classification score and delay. Also an input window of 10 time steps is quite prone to false positives, since there is often movement while waiting, e.g., switching the leg on which the cyclist rests, preparing the pedals or simply looking around. Due to the same reasons filter features are preferred to the raw data , because spikes of activity are smoothed out over the feature windows. An overview of the delay distribution can be seen in Fig. \ref{fig:filter_06_results}. Most starting detection occur between 0.5\,s and 1\,s after the \textit{moving} label. It is noticeable though, that some detections happen almost 1\,s before the bicycle moves.

The bigger models did not reach the same performance on the test set, having a maximal $F_1$-score of 93\% with a keep probability of 0.8. This is supposedly because of over-fitting since the training loss was comparable to the smaller models. Although, the mean delay times improved with the model size, reaching a minimal mean delay of 592\,ms with a keep probability of 0.7. A decisive conclusion on the keep probability could not be drawn due to no apparent trend for it being high or low.

Choosing a minimum desired $F_1$-score of 95\%, the best model was achieved by the small network and a keep probability of 0.6. It reached a mean $F_1$-score of 97\% and delay of 783\,ms.

\section{Conclusion}
\label{js:sec:Conclusion}
The high scores on the selected models show, that the use of convolutional neural network on sensory data from smart phones is justified. Our approach reaches an $F_1$-score of 97\% with an average delay of 783\,ms. Deep Residual Neural Networks  proved to be applicable for the task of starting detection. They achieve a reliable classification, although with a comparatively high delay of over half a second. 
Further improvements can be achieved with more training data.
Cutting the data to not include stopping motion greatly improved scores, since very high excitations are basically only encountered at the starting motion. This could be implemented with a separate waiting detector, which activates the starting detector only when the stopping motion was finished and switching back to it once the driving.

\section*{\large Acknowledgment}

This work results from the project DeCoInt$^2$, supported by the German Research Foundation (DFG) within the priority program SPP 1835: "Kooperativ interagierende Automobile", grant number SI 674/11-1.

\bibliographystyle{splncs}
\bibliography{js_references}

\begin{thebibliography}{10}

\bibitem{mbBRZ+17}
Bieshaar, M., Reitberger, G., Zernetsch, S., Sick, B., Fuchs, E., Doll, K.:
\newblock Detecting intentions of vulnerable road users based on collective
  intelligence.
\newblock In: AAET – Automatisiertes und vernetztes Fahren, Braunschweig,
  Germany (2017)  67--87

\bibitem{mbBZD+17}
Bieshaar, M., Zernetsch, S., Depping, M., Sick, B., Doll, K.:
\newblock Cooperative starting intention detection of cyclists based on smart
  devices and infrastructure.
\newblock In: ITSC, Yokohama, Yapan (2017)

\bibitem{singh2017}
Singh, M.S., Pondenkandath, V., Zhou, B., Lukowicz, P., Liwicki, M.:
\newblock Transforming {S}ensor {D}ata to the {I}mage {D}omain for {D}eep
  {L}earning - an {A}pplication to {F}ootstep {D}etection.
\newblock (2017)

\bibitem{jiang2015}
Jiang, W., Yin, Z.:
\newblock Human {A}ctivity {R}ecognition using {W}earable {S}ensors by {D}eep
  {C}onvolutional {N}eural {N}etworks.
\newblock In: Proceedings of the 23rd ACM international conference on
  Multimedia. (2015)  1307--1310

\bibitem{mbBZH+18}
Bieshaar, M., Zernetsch, S., Hubert, A., Sick, B., Doll, K.:
\newblock Cooperative starting movement detection of cyclists using
  convolutional neural networks and a boosted stacking ensemble.
\newblock CoRR \textbf{arXiv:1803.03487} (2018)

\bibitem{zernetsch2018}
Zernetsch, S., Kress, V., Sick, B., Doll, K.:
\newblock Early {S}tart {I}ntention {D}etection of {C}yclists {U}sing {M}otion
  {H}istory {I}mages and a {D}eep {R}esidual {N}etwork.
\newblock (2018)

\bibitem{zernetsch2018_2}
Zernetsch, S., Kohnen, S., Goldhammer, M., Doll, K., Sick, B.:
\newblock Trajectory {P}rediction of {C}yclists {U}sing a {P}hysical {M}odel
  and an {A}rtificial {N}eural {N}etwork.
\newblock In: Intelligent Vehicles Symposium (IV), IEEE. (2016)

\bibitem{Bulling2014THA}
Bulling, A., Blanke, U., Schiele, B.:
\newblock A tutorial on human activity recognition using body-worn inertial
  sensors.
\newblock ACM Computing Surveys \textbf{46} (2014)  1--33

\bibitem{ILSVRC15}
Russakovsky, O., Deng, J., Su, H., Krause, J., Satheesh, S., Ma, S., Huang, Z.,
  Karpathy, A., Khosla, A., Bernstein, M., Berg, A.C., Fei-Fei, L.:
\newblock {ImageNet Large Scale Visual Recognition Challenge}.
\newblock International Journal of Computer Vision (IJCV) \textbf{115} (2015)
  211--252

\bibitem{he2015}
He, K., Zhang, X., Ren, S., Sun, J.:
\newblock Deep {R}esidual {L}earning for {I}mage {R}ecognition.
\newblock CoRR \textbf{arXiv:1512.03385v1} (2015)

\bibitem{ioffe2015}
Ioffe, S., Szegedy, C.:
\newblock Batch {N}ormalization: {A}ccelerating {D}eep {N}etwork {T}raining by
  {R}educing {I}nternal {C}ovariate {S}hift.
\newblock In: ICML'15 Proceedings of the 32nd International Conference on
  International Conference on Machine Learning. Volume~37. (2015)  448--456

\bibitem{he2016}
He, K., Zhang, X., Ren, S., , Sun, J.:
\newblock Identity {M}appings in {D}eep {R}esidual {N}etworks.
\newblock CoRR \textbf{arXiv:1603.05027v3} (2016)

\bibitem{ng2017}
Ng, A.:
\newblock Why does {B}atch {N}orm {W}ork? (c2w3l06) (2017) Deeplearning.ai
  (Last accessed: 28.05.2018).

\end{thebibliography}

\end{document}